\documentclass{article}
\usepackage[utf8]{inputenc}
\usepackage[T5]{fontenc}
\usepackage[preprint]{corl_2026} 
\usepackage{graphicx}
\usepackage{booktabs}
\usepackage{subcaption}
\usepackage{multirow}
\usepackage{float}
\usepackage{array}
\usepackage{makecell}
\usepackage[table,xcdraw]{xcolor}
\usepackage{colortbl}
\usepackage{makecell}
\usepackage[colorinlistoftodos, textsize=small]{todonotes}
\usepackage[ruled,vlined]{algorithm2e}
\usepackage{float}
\usepackage{wrapfig}
\usepackage{amsmath}
\usepackage{amssymb}
\usepackage{wrapfig}
\usepackage{placeins} 
\usepackage[accsupp]{axessibility}

\usepackage{adjustbox}
\usepackage{makecell}

\title{Robot Critics that Sweat the Small Stuff}

\author{
\textbf{Sruthi Sudhakar}$^{1}$ \quad
\textbf{Junbang Liang}$^{1}$ \quad
\textbf{Sreehari Rammohan}$^{1}$\\
\textbf{Pavel Tokmakov}$^{2}$ \quad
\textbf{Richard Zemel}$^{1}$ \quad
\textbf{Carl Vondrick}$^{1}$\\[0.5em]
$^{1}$Columbia University \quad
$^{2}$Toyota Research Institute\\[0.3em]
\href{https://robocritic.cs.columbia.edu/}{\url{robocritic.cs.columbia.edu}}
}
\hypersetup{
  pdftitle={Robot Critics that Sweat the Small Stuff},
  pdfauthor={Sruthi Sudhakar, Junbang Liang, Sreehari Rammohan, Pavel Tokmakov, Richard Zemel, Carl Vondrick},
  pdfkeywords={robotics, vision-language models, behavior critics}
}
\begin{document}
\maketitle

\begin{abstract}
Large vision-language models contain several priors about the world and object interactions, making them useful critics during inference to steer robot policies towards success. However, closed-loop robot manipulation requires judging small visual differences between success and failure, which remains a challenge for current VLMs. We introduce a method to fine-tune critics by constructing pairwise progress supervision using success and failure rollouts obtained from a policy. Our fine-tuned critic excels at fine-grained progress reasoning and subtle failure detection, outperforming prior progress reasoning baselines. Additionally, we use an action-conditioned video model to predict the visual effect of several candidate actions sampled from a policy, and show that our critic can correctly identify successful candidates to execute, improving the average policy success rate by 11\% across real-world tasks and 5.9\% across simulation tasks.
\end{abstract}

\keywords{Vision Language Models, Imitation Learning, Video Generation} 

\section{Introduction}
\label{sec:intro}

Humans rely on strong visual reasoning to decide which actions will complete a task and which will lead to failure. Recent advances in vision-language foundation models (VLMs)~\citep{alayrac2022flamingo, liu2023visual, driess2023palm} show that these models encode extensive knowledge about the physical world, providing hope of using them for robot policies. One promising approach is to use VLMs as critics in the policy execution loop to reason over multiple candidate actions sampled from a policy~\citep{Wu2025FromFT,Zhao2024VLMPCVM}. The critic can compare the candidate trajectories and select the action that makes the greatest progress towards success, thereby steering away from failures. Such progress/failure-aware critics aid standard behavior cloning policies which cannot reason about failures because they have only been trained on successful demonstrations.

We find that, while off-the-shelf VLMs~\citep{Bai2025Qwen25VLTR, Comanici2025Gemini2P} and progress detection methods~\citep{Schroeder2025ROVERRR,Ma2024VisionLM} can distinguish initial from final task states and identify high-level subtasks, they fail to discriminate between fine-grained progress and failures. A subtle 1mm gripper misalignment is close, but still insufficient, causing the overall policy to fail. On eight robot tasks, Gemini 3 discriminates success versus failure with only 54.7\% accuracy, and  the progress detection methods ROVER and ProgressLM~\citep{Schroeder2025ROVERRR,Zhang2026PROGRESSLMTP} achieve only 72.1\% and 61.0\%  accuracy respectively -- where chance is 50\%. ROVER achieves much higher accuracy on coarse progress detection, highlighting that these methods struggle in the fine-grained regime where the time interval gap narrows. 

\begin{figure}
    \centering
    \includegraphics[width=0.9\linewidth]{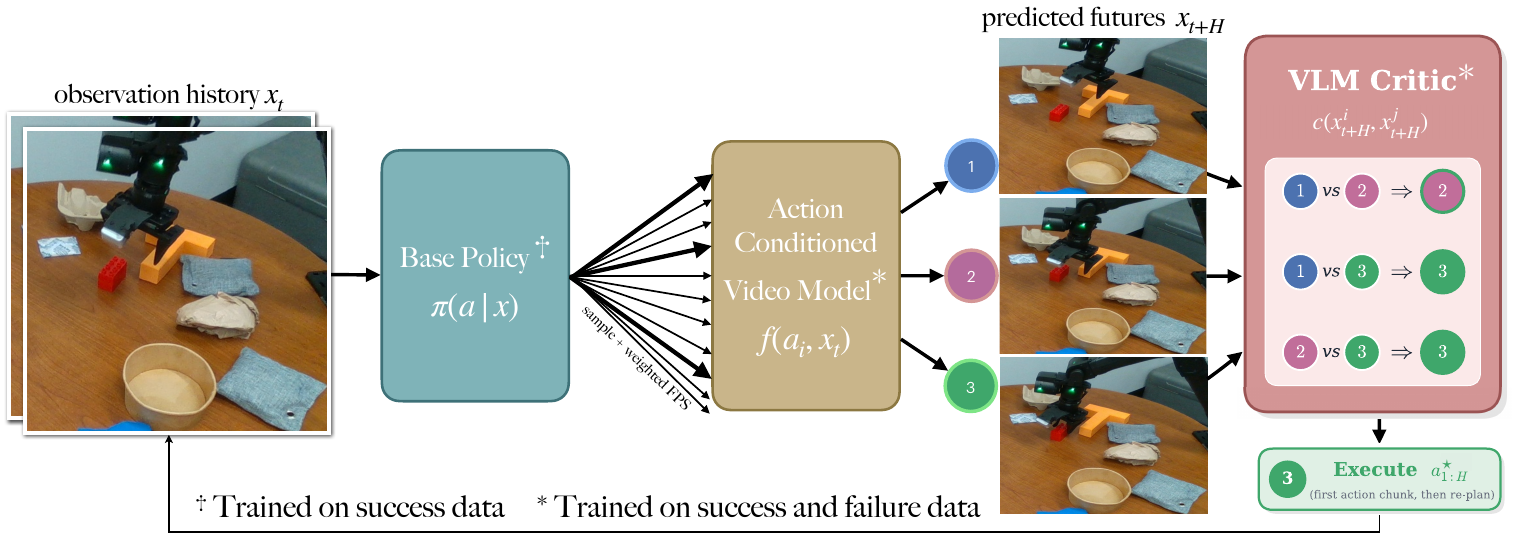}
    \caption{\textbf{Critic in-the-loop.} Given an observation, a learned stochastic policy samples $K$ unique candidate action sequences. An action-conditioned generative video model synthesizes visual observations to produce a terminal state per candidate action. The critic performs pairwise progress comparisons to select the best candidate, which is then executed and the re-planning continues.}
    \label{fig:policyintheloop}
    \vspace{-0.2in}
\end{figure}

We introduce a framework to fine-tune VLMs for fine-grained progress/failure discrimination, and use them as critics to improve robot policy performance in-the-loop. To fine-tune VLMs, we generate paired success and failure rollouts from matched initial conditions, enabling construction of a fine-grained progress comparison dataset. We emphasize the need to incorporate policy specific failures to train the critic. This procedure only requires binary success/failure labels at the end of the trajectory, and naturally captures the failure modes of the deployed policy. We then integrate the critic into a robot policy for test-time performance improvement via policy steering. We sample multiple candidate action sequences from the policy, use a generative video model to synthesize future visual states, and select which one to execute using the critic (Fig.~\ref{fig:policyintheloop}). 

Critics fine-tuned with this framework achieve over 90\% accuracy and substantially outperform baselines in both simulation and real world experiments, and generalize to tasks unseen during training. We systematically evaluate the critic in-the-loop across real world tasks and simulation tasks. We see an average +11\% improvement in success rate and 2.3x improvement in mean intersection over union (mIoU) over the base policy on real world tasks, and +5.9\% improvement in success rate across all simulation tasks.

\section{Related Work}

\textbf{Robot policies and test-time policy steering.}
Modern robot manipulation policies include both task-specific visuomotor policies and large vision-language-action (VLA) models that can model multimodal action distributions~\citep{Chi2024UniversalMI,Khazatsky2024DROIDAL,Padalkar2023OpenXR,Pumacay2024THECA,KressGazit2024RobotLA}. Recent VLA and generalist robot policies further scale this paradigm with larger backbones and action chunking~\citep{Intelligence202505AV,black2025real,wu2026pragmatic,xie2026dynamicvla,beyer2024paligemma, liang2024dreamitate,liang2025videogeneratorsrobotpolicies}. Many of these policies share an important test-time property: stochastic policies can produce multiple plausible action candidates from the same observation. This creates an opportunity for policy improvement without retraining. Several methods improve policies at test time by steering~\citep{Wang2024InferenceTimePS,Du2025DynaGuideSD,nakamoto2025steeringgeneralistsimprovingrobotic}, reranking or sampling candidate actions~\citep{jang2025verifier,guo2025vla}, or composing policies~\citep{cao2025compose}. Related work also uses predictive world models to strengthen or guide robot policies~\citep{Qi2025StrengtheningGR,gao2026dreamdojo}. These approaches show that predicting future visual states is effective in robot learning. 

\textbf{Vision-language models as critics.}
Large vision-language models have increasingly been used in robotics as planners, reward functions, value estimators, and critics~\citep{driess2023palm,Liang2022CodeAP,Gao2023PhysicallyGV,Ma2023LIVLR,Du2023VisionLanguageMA,Zhou2025PhysVLMEV,Zhai2025AVM}. For example, SayCan grounds language-model plans in robotic affordances to select feasible skills~\citep{Ahn2022DoAI}. VLMs are also used as feedback or scoring functions for either training policies with RL~\citep{Wang2024RLVLMFRL} or optimizing actions in a model predictive control loop~\citep{finn2017deep}. In VLMPC, a VLM-derived hierarchical cost function is used within model-predictive control to evaluate candidate action sequences~\citep{Zhao2024VLMPCVM}. Recent VLM-in-the-loop steering methods use pretrained VLM representations or latent alignment objectives to guide action selection~\citep{Wu2025FromFT}. These methods show that VLMs can provide useful semantic priors for robot decision-making. While these works rely on relatively coarse judgments, closed-loop policy steering towards success requires differentiating between small changes in gripper pose, object contact, alignment, or early signs of failure. Our work therefore studies VLMs in a fine-grained pairwise comparison between policy-action candidates, rather than as generic semantic goal verifiers/reward functions.

\textbf{Progress reasoning and failure detection.}
Several methods aim to estimate task progress from visual observations~\citep{Xue2024ProgressAwareVF,Hung2024VICtoRLH,Pacaud2025GuardianDR,Agrawal2015VQAVQ,Hudson2019GQAAN,Yue2023MMMUAM}. Generative Value Learning (GVL) treats VLMs as in-context value learners to obtain the progress value of a new observation~\citep{Ma2024VisionLM}. ROVER decomposes embodied videos into subtasks and uses recursive VLM reasoning to estimate task progress~\citep{Schroeder2025ROVERRR}. ProgressLM fine-tunes a VLM to predict absolute progress scores for frames given task demonstrations or key frames~\citep{Zhang2026PROGRESSLMTP}. RoboMeter~\citep{Liang2026RobometerSG} is a reward model to learn general failure cases using failed/suboptimal trajectories, whereas we focus on policy-specific failures and fine-grained differences. Recent work has studied how to classify and use failures at inference time for recovery ~\citep{xu2025detectfailuresfailuredata,Duan2024AHAAV,lin2025failsafe,park2025hierarchicalvisionlanguageaction,li2026failureawarerlreliableofflinetoonline}. These methods are closely related because they also ask whether VLMs can reason about temporal progress in embodied tasks, but they typically estimate more coarse progress on individual frames or videos. Our critic is related to preference-learning and reward-modeling approaches, where a model is trained from comparisons rather than absolute labels~\citep{christiano2023deepreinforcementlearninghuman,bradley1952rank}. 
We focus on learning fine-grained pairwise progress critics from successful and failure policy rollouts. This lets the critic recognize both task progress and policy-specific failure modes, which can then be avoided during test-time selection.

\section{Method}
\label{sec:method}

We use VLMs as critics to guide stochastic robot policies towards success at test-time. Given an image $x$, the policy predicts absolute joint-angle actions $a^* \in \mathbb{R}^{D \times H}$ to complete the task, where $D$ is the number of degrees of freedom and $H$ is the policy horizon. Our approach will sample multiple action trajectories from a base policy $\pi(x)$ and use a critic to select the final action trajectory:
\begin{align}
a^* = \arg\min_{a_i \sim \pi(x)} \sum_{j \neq i} c(v_i, v_j) \quad \textrm{where} \quad v_i = f(a_i, x).
\end{align}
$f$ is a generative video model producing the final frame $v$ which acts as a forward dynamics model, and $c$ is a critic that scores the utility of a rollout (Fig~\ref{fig:policyintheloop}). We instantiate base policy $\pi$ as any stochastic robot policy (diffusion policy or GR00T N1.5 in our experiments), $f$ as a video diffusion model, $c$ as a large vision-language model. 

We start with our behavior cloned base policy that is already trained on expert demonstrations, and attach a critic to it. We show that we can design this critic to steer the base policy towards successful outcomes, allowing us to integrate the knowledge from VLMs into policies.

\subsection{Training VLMs as Critics}
\label{sec:training_vlms_as_critics}

We train a VLM to predict which frame reflects greater progress toward the task goal. Note that lack of progress can also mean failure has occurred. Given two frames $(v_i, v_j)$ from a robot manipulation trajectory and a natural language description of the task $\mathcal{T}$, the VLM makes a discrete decision $c(v_i, v_j) \in \{+1,-1\}$ by outputting a text token (preserving the VLM's natural output space). 
A $+1$ indicates $v_j$ shows more progress than $v_i$, and $-1$ indicates the opposite. We provide $v_i$ and $v_j$ to the VLM as two separate images alongside the task description in context. We do supervised finetuning on our dataset $D$ to minimize the cross entropy loss:
\begin{equation}
\label{eq:critic_loss}
\mathcal{L}_{\text{critic}}(\theta)
=
-\frac{1}{|D|}
\sum_{(v_i,v_j,y)\in D}
\log c_\theta\!\left(y \mid v_i, v_j\right)
\quad \textrm{for} \quad
y \in \{+1,-1\}.
\end{equation}

\begin{figure}
    \centering\includegraphics[width=0.81\linewidth]{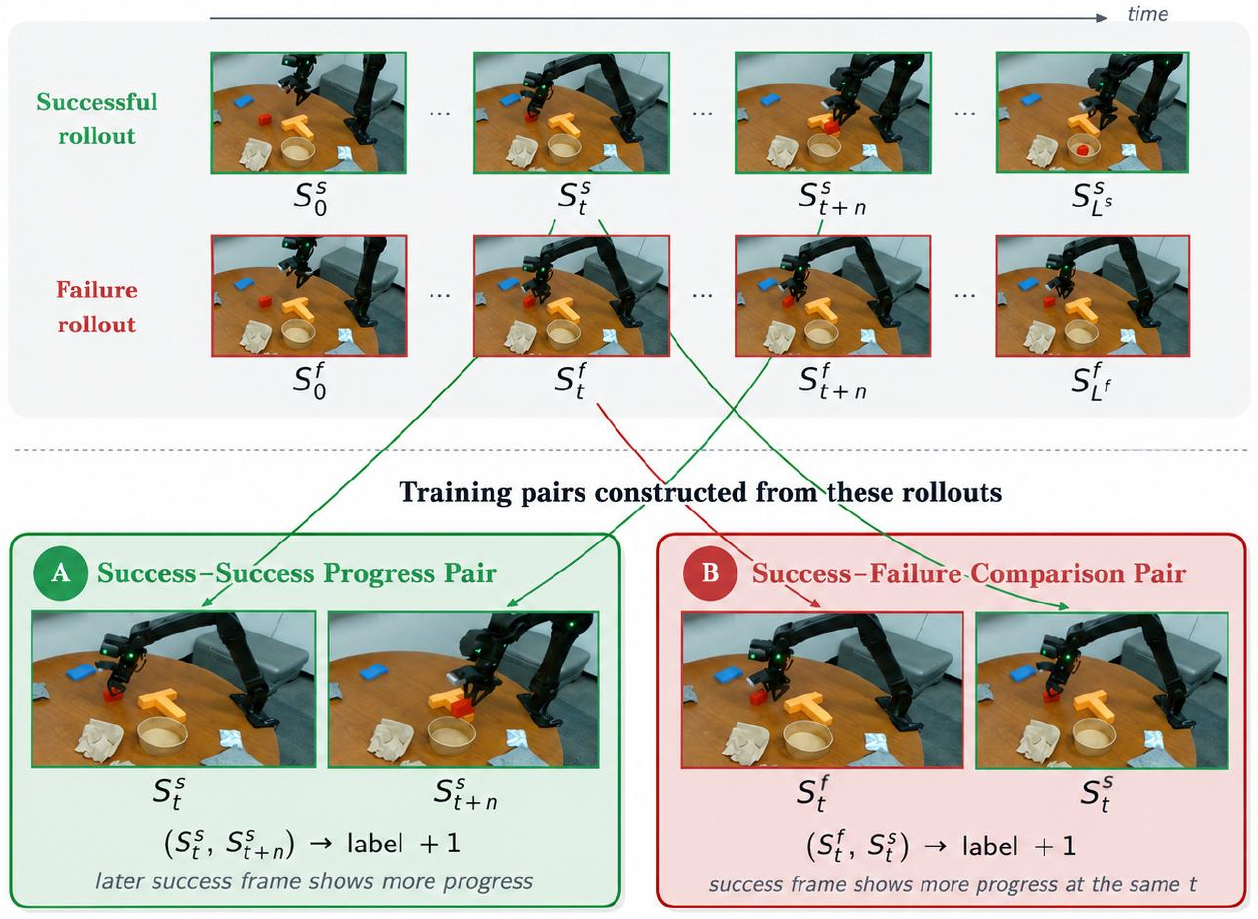}
    \caption{\textbf{Training Robot Critics.} VLMs are finetuned with successful and failed rollouts to enable fine-grained task progress/failure detection. To construct the training dataset, consecutive frames are used from successful trajectories to obtain task progress data. Additionally, success and failure frame pairs from the same initial condition are used to obtain policy- and task-specific failure data.}
    \label{fig:vlmmethod}
    \vspace{-10pt}
\end{figure}

To construct a supervised fine-tuning dataset $D$ tailored to learning fine-grained visual task progress/failure, we roll out the policy on the same set of initial conditions multiple times
to obtain paired success $S^s$ and failure trajectories $S^f$ (top half of Fig \ref{fig:vlmmethod}). In cases where recording paired failures from the policy is infeasible, we note that human demonstrations of failure trajectories can be quickly and easily collected.
We construct pairs of frames from $S^s$ and $S^f$:

\textbf{Success-Success Progress Pairs.} For each successful trajectory $S^s$ of length $L^s$, and for each temporal interval $l \in \{4,8,12,16\}$, a pair is defined as consecutive observations from an earlier and later timestep in the trajectory, 
$
(S^s_{t}, S^s_{(t+l)}) $ for $t \in \{0, \ldots, L^s - l\}
$. Under successful executions, later timesteps typically reflect greater completion toward the goal; therefore, $S^s_{(t+l)}$ reflects greater completion than $S^s_{t}$ and the pair is assigned label $+1$ (as visualized in Figure \ref{fig:vlmmethod}, bottom left).
These success-success pairs teach the model to recognize gradual forward progress.

\textbf{Success–Failure Comparison Pairs.} We observe that fine-tuning the VLMs on the successful demonstrations is insufficient as they cannot identify failures without ever having seen failure data.
To address this limitation, we include failure trajectory rollouts, $S^f$, that originate from the same initial conditions as a successful rollout, $S^s$. We construct pairs $(S^f_{t}, S^s_{t})$ for $ t \in \{0, \ldots, L^f\}$, and label them $+1$ as visualized in bottom right box in Fig~\ref{fig:vlmmethod} (i.e., the success frame shows more task progress than the failure frame at the same point in time). This teaches the critic what ``bad actions'' look like, tailored to this specific policy, task, and environment combination. Creating success-failure pairs with the same initial conditions is important as it helps the VLM learn what kinds of actions cause subtle failures versus what are simply variances in object/robot pose. 

\subsection{Integrating Critics into the Policy}

We use the fine-tuned critic in the loop for test-time policy improvements. At each re-planning step, given the observation $x_t$, we draw $N$ i.i.d. candidate action chunks from the base policy: $a^{(i)} \sim \pi(x_t)$ for $i \in \{1,\dots,N\}$. We do greedy farthest-point sampling (FPS) over a weighted joint-space L2 distance to prune to a diverse subset of $K$ actions. Each candidate $a^{(i)}$ specifies a sequence of low-level actions; to derive images that the critic can compare between, we execute all $K$ action chunks with an action-conditioned video model to obtain the terminal states $x_{t+H}^{(i)}=f(a_i,x_t)$. 

Action-conditioned video prediction models allow a robot to learn how its actions affect objects in its environment~\citep{Finn2016UnsupervisedLF, yang2024learninginteractiverealworldsimulators}. We use the HunyuanVideo 1.5 pretrained image-to-video Diffusion Transformer~\cite{wu2025hunyuanvideo15technicalreport}, a state-of-the-art video generative model. To condition on a sequence of actions, we insert cross-attention blocks that attend to the input action sequence $a$~\cite{li2025multimodalactionconditionedvideo}. We encode the sequence using a transformer encoder to produce a single token per frame, and fine-tune the model using a flow-matching objective to generate videos conditioned on the encoded action tokens. Each action input corresponds to one frame of the predicted video. More details provided in Section \ref{sec:video_model_details}.

We then form all $\binom{K}{2}$ pairwise comparisons of these terminal states and query our learned critic to determine which state reflects greater task progress. Each comparison awards a vote to the preferred candidate. We select the candidate whose terminal state accumulates the most votes. The winning action $a^*$ is then executed in the real environment and this sample-and-select procedure is applied at every re-planning step until task completion or failure.
\section{Critic Experiments}
\label{sec:exp}
We aim to understand the performance of policy critics by answering the following key questions: \textbf{Q1.} Does our finetuning approach enable stronger critics in different tasks and environments? \textbf{Q2.} Is failure data important? \textbf{Q3.} How does our method compare on increasingly fine-grained detection? 

\subsection{Robot Dataset Setup}
\label{sec:robot_dataset_setup}
\textbf{RoboCasa.} We use RoboCasa~\citep{Nasiriany2024RoboCasaLS} to evaluate task-progress assessment. 
To generate both success and failure data, we train a policy $\pi$ on each task using the provided 50 human demonstrations. We then roll out the policy $\pi$ multiple times and collect successful and failure trajectories per initial condition for each task. Across 30 initial conditions, we collect paired success and failure trajectories, and we construct the training dataset as described in Section~\ref{sec:training_vlms_as_critics} which generates approximately 21k pairs per task. We hold out 7 success and 7 failure trajectories for evaluation (about 5k pairs). We report accuracy on \textbf{ID} (in-distribution objects) and \textbf{OOD} (out-of-distribution objects).

\textbf{LBM.} We also evaluate our critic finetuning method on the LBM dataset~\citep{Team2025ACE}. We use a publicly released portion of the \cite{Team2025ACE} dataset which has 8 tasks including: BimanualBikeRotorInstall, BimanualClearKitchenCounter, BimanualSetUpBreakfastTable, CleanLitterBox, CutAppleIntoSlices, PushCoasterToMug, PutKiwiInCenterOfTable, and TurnMugRightsideUp. For each dataset, there are 50 initial conditions that 2 or 3 policies are rolled out on and videos are generated from: either a Single Task policy, the Finetuned LBM policy, and for some tasks there additionally a rollout from the pre-trained LBM policy (policy trainings are described in \cite{Team2025ACE}). These rollouts lead to either successes or failures which are labeled by human annotators. We create two splits: 40 initial conditions per task for fine-tuning and 10 held-out initial conditions for testing.

\subsection{Baselines}
We compare against 3 types of baselines: out-of-the-box foundation models, prompted progress detection methods, and fine-tuned progress detection methods. See Section \ref{sec:prompt_details} for prompts.

\textbf{Out-of-the-box foundation models.} We evaluate pretrained foundation models including \textbf{Qwen2.5-VL}~\citep{Bai2025Qwen25VLTR} and \textbf{Gemini 3}~\cite{Comanici2025Gemini2P} without finetuning. Similar to our method, we prompt these VLMs with the image pair along with the robot task/environment descriptions and the discrimination task description. These methods output a binary label $y\in\{+1,-1\}$. 

\textbf{Prompted progress detection.} We evaluate two prompted progress-reasoning methods. \textbf{GVL}~\citep{Ma2024VisionLM} predicts per-frame progress given in-context progress annotations along a demonstration trajectory. \textbf{ROVER}~\citep{Schroeder2025ROVERRR} decomposes videos into subtasks using recursive prompting, predicting subtask id and progress scores. For both methods, we compute the signed difference between each frame's progress scores to obtain binary labels $y\in\{+1,-1\}$.

\textbf{Fine-tuned progress detection.} We also compare to \textbf{ProgressLM}~\citep{Zhang2026PROGRESSLMTP}, which fine-tunes Qwen2.5-VL to predict an absolute progress score from demonstration key frames with known progress percentages. We compute the signed difference between the absolute score of the two images in the pair to obtain binary labels $y\in\{+1,-1\}$. We fine-tune ProgressLM-7B on the same RoboCasa dataset, using 10 demonstration key frames per trajectory. 

\textbf{Ours.} We fine-tune Qwen2.5-VL-7B with Eq.~\eqref{eq:critic_loss} on successful and failure rollouts. We also run an ablation where we fine-tune only on successful rollouts.

\begin{table*}[t]
\centering
\scriptsize
\setlength{\tabcolsep}{3.8pt}
\renewcommand{\arraystretch}{1.08}
\begin{tabular}{l c c c c c c c c c}
\toprule
\textbf{Method}
& \makecell{\textbf{Countr}\\\textbf{Cab}}
& \makecell{\textbf{Cab}\\\textbf{Countr}}
& \makecell{\textbf{Cab}\\\textbf{Micro}}
& \makecell{\textbf{Micro}\\\textbf{Cab}}
& \makecell{\textbf{Countr}\\\textbf{Sink}}
& \makecell{\textbf{Sink}\\\textbf{Countr}}
& \makecell{\textbf{Countr}\\\textbf{Stove}}
& \makecell{\textbf{Stove}\\\textbf{Countr}}
& \textbf{Avg.} \\
\midrule
\textbf{Overall Accuracy} \\
Qwen2.5-32B
& 0.495 & 0.508 & 0.502 & 0.489 & 0.502 & 0.491 & 0.512 & 0.490 & 0.499 \\
Gemini 3
& 0.706 & 0.578 & 0.428 & 0.498 & 0.659 & 0.681 & 0.398 & 0.424 & 0.547 \\
GVL
& 0.528 & 0.462 & 0.584 & 0.527 & 0.576 & 0.543 & 0.538 & 0.486 & 0.531 \\
ROVER
& 0.707 & 0.677 & 0.798 & 0.758 & 0.727 & 0.657 & 0.710 & 0.737 & 0.721 \\
ProgressLM
& 0.206 & 0.498 & 0.709 & 0.533 & 0.850 & 0.597 & 0.774 & 0.715 & 0.610 \\
Ours (Success-Only)
& 0.927 & 0.876 & 0.874 & 0.892 & 0.892 & 0.858 & 0.886 & 0.885 & 0.886 \\
{Ours (Full Method)}
& \textbf{0.966} & \textbf{0.931} & \textbf{0.908} & \textbf{0.942}
& \textbf{0.965} & \textbf{0.918} & \textbf{0.916} & \textbf{0.943}
& \textbf{0.936} \\
\midrule
\textbf{OOD Only Accuracy}\\
ProgressLM
& 0.200 & 0.495 & 0.704 & 0.598 & 0.843 & 0.502 & 0.770 & 0.680 & 0.599 \\
Ours (Success-Only)
& 0.940 & 0.851 & 0.829 & 0.872 & 0.894 & 0.792 & 0.897 & 0.896 & 0.871 \\
{Ours (Full Method)}
& \textbf{0.970} & \textbf{0.909} & \textbf{0.863} & \textbf{0.916}
& \textbf{0.958} & \textbf{0.870} & \textbf{0.903} & \textbf{0.924}
& \textbf{0.914} \\
\bottomrule
\end{tabular}%
\caption{\textbf{Critic accuracy} on various tasks from the RoboCasa dataset~\cite{Nasiriany2024RoboCasaLS}. In-distribution (ID) refers to objects
the VLM has seen during training. Out-of-distribution (OOD) are unseen objects on the same tasks. We report average in-distribution and out-of-distribution accuracy. For fine-tuned methods, we also report OOD accuracy separately. Furthermore, we show that finetuning with success and failure data (Ours) leads to stronger fine-grained progress/failure understanding than just success (Ours Success-only).}
\label{tab:table1}
\vspace{-10pt}
\end{table*}

\subsection{Results}

\textbf{Q1. Observing subtle failures creates a strong critic.} Out-of-the-box models have never seen the nuances of how a specific policy can fail, something that is typically only understood after observing several rollouts of success and failures. Consistent with this observation, Tables~\ref{tab:table1} and~\ref{tab:lbm_details} show that prompted, off-the-shelf VLMs are unreliable for fine-grained task progress. Our method shows strong ID and OOD accuracy on real and simulation datasets. Furthermore, our method generalizes to new tasks that it was never fine-tuned on, suggesting a promising pretraining strategy for future task learning (full results in Section \ref{sec:gen_unseen_tasks}).
Our fine-tuned critic learns to both de-prioritize actions that move away from the goal (or failures) and prefer states that are closer to task completion. This fine-grained capability is crucial for using critics as practical guidance signals in closed-loop policy selection as we investigate in Section~\ref{sec:realworld_analysis}. 

\begin{wraptable}[10]{r}{0pt}
\vspace{-0.5in}
\centering
\tiny
\setlength{\tabcolsep}{2.5pt}
\renewcommand{\arraystretch}{0.78}
\begin{tabular}{@{}lcccc@{}}
\toprule
Task & Gem. & ROVER & ProgLM & Ours \\
\midrule
BimanBike   & .473 & .717 & .443 & \textbf{.980} \\
BimanClear  & .492 & .545 & .709 & \textbf{.988} \\
BimanSetUp  & .545 & .758 & .613 & \textbf{.889} \\
CleanBox    & .510 & .615 & .615 & \textbf{1.00} \\
CutApple    & .697 & .646 & .661 & \textbf{.919} \\
PushCoaster & .608 & .712 & .667 & \textbf{.885} \\
PutOnTable  & .510 & .490 & .615 & \textbf{.778} \\
TurnUpright & .596 & .547 & .600 & \textbf{.763} \\
\textbf{Avg.} & .554 & .629 & .615 & \textbf{.900} \\
\bottomrule
\end{tabular}

\vspace{-0.08in}
\caption{\textbf{Critic accuracy} on the real world LBM dataset.}
\label{tab:lbm_details}
\vspace{-0.16in}
\end{wraptable}
\textbf{Q2. Failure data is necessary for useful critics.}
Our results show that finetuning on only successful rollouts does not teach the critic to recognize failure modes. Critics finetuned only on success leads to a 26\% drop in accuracy on success-failure pairs, whereas a critic fine-tuned on both success and failure rollouts yields strong performance on both success-success and success-failure pairs. One explanation for this might be that despite foundation VLMs' internet knowledge, many robot datasets that exist on the internet are often released for fine-tuning purposes. Therefore, these models may have a strong bias towards videos of robots succeeding at a task. This means common failure modes are under-represented in large pretrained foundation models. Furthermore, generic failure modes are hard to quantify, and policy-specific failures can look very different. 

\textbf{Q3. Fine-grained Capabilities Comparison.} 
To isolate the effect of visual difference magnitude, we evaluate ROVER, ProgressLM, and our method under two sampling regimes. In the \emph{fine-grained} regime, we sample query frames with small temporal intervals ($l=16, 32, 48$ steps $\approx 1-5s$ apart), where differences are subtle. In the \emph{coarse} regime, we sample frames at larger intervals at more than seven seconds apart, where visual differences are more pronounced. As the frame interval length decreases, performance for prior methods drops from an average of 73\% to 56\% across the eight RoboCasa tasks, indicating that these models are more reliable when differences are large (see plot in \ref{sec:finevscoarse}). Practical policy guidance, however, requires performance in the fine-grained regime as well, a regime where our critic excels with only a 2\% drop.

\section{Critic In-The-Loop Experiments}
\label{sec:realworld_analysis}

Here we test whether these critics translate into closed-loop robot policy improvements in both real-world (Sec~\ref{sec:realworld_results}) and simulation (Sec~\ref{sec:sim_policy_results}) settings. 

To understand the upper bound on our approach, we consider an idealized case, where we had an oracle critic that always picked the best rollout sampled. This improved the policy performance from 31\% to 48\% (see Section \ref{sec:accvsk_results} for acc-vs-K plot), showing that the base policy does assign significant probability density to successful rollouts, but does not always sample them. This result suggests that, if we can learn a strong critic, we can potentially improve the policy performance substantially. 

\subsection{Real World Experiments}
\label{sec:realworld_results}

\textbf{Setup.} We experiment on tabletop manipulation tasks using 7-DoF YAM Pro Arms from I2RT. The workspace is observed by two Intel RealSense D435i cameras placed approximately $660$ mm apart and angled toward the table at roughly $45^\circ$. We use the Yam Pro Leader arm to teleoperate the robot to collect data for all real world tasks (Appendix Table \ref{tab:realworld_dataset_stats} details the number of multi modal expert demonstrations we collect per
task). Random demonstrations show the expert randomly moving in the environment and interacting
with objects. Only successful demos are used for training the policy, successful and failure demos
are used for finetuning the VLM critic, and finally successful, failure, and random demos are used
for finetuning the world model. The breakdown of the finetuning dataset for the VLM critic is shown
in Appendix Table \ref{tab:vlm_dataset_details}.

\begin{figure}
    \centering
    \includegraphics[width=\linewidth]{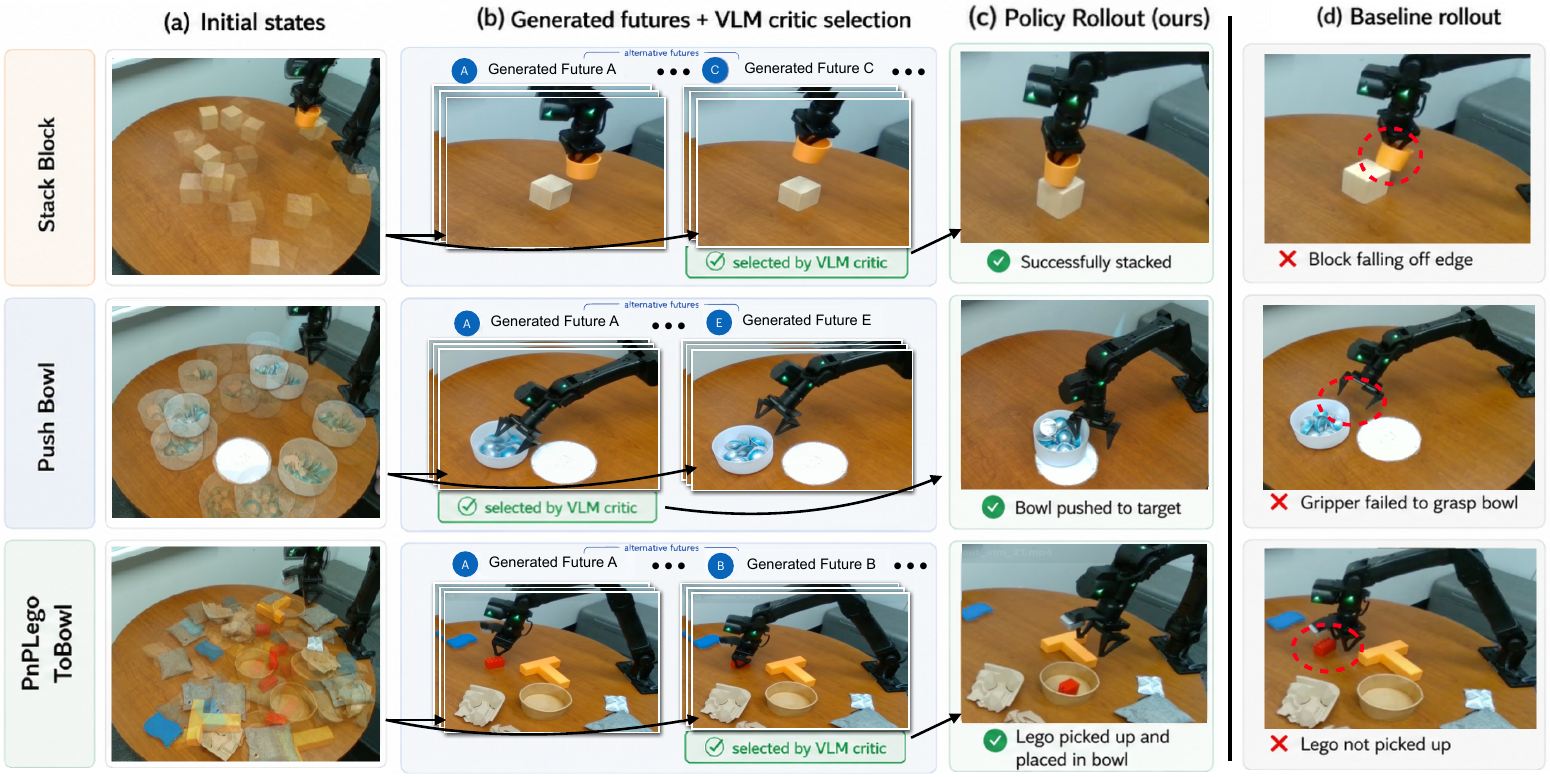}
    \caption{\textbf{Real-world Evaluation Results}. (a) Initial states for all eval episodes overlaid - all methods start with the same set of initial states, matched manually with reference images. (b) Critical decision points where the VLM chooses the correct action candidate to lead to success (c) Our method succeeds at the task at the end of the rollout (d) The baseline policy fails at those critical points and therefore fails to complete the task.}
    \label{fig:realworld_qual_results}
    \vspace{-10pt}
\end{figure}

We evaluate on multiple tasks:
(i) Stacking a cup onto a block, (ii) Pushing a bowl onto a white place mat, (iii) Pick and Place a Lego block into a brown bowl in the presence of eight distractor objects, and (iv) Pickup Lego. For each task, we collect expert teleoperated demonstrations at $30$ Hz. Details on number of demonstrations per task and model included in Table \ref{tab:realworld_dataset_stats}. All real-world results are evaluated over $50$ trials per task with varying unseen initial conditions as seen in Fig~\ref{fig:realworld_qual_results}a.

Our base policy $\pi$ is a CNN-based Diffusion Policy~\citep{Chi2023DiffusionPV}. We subsample expert trajectories at 10 Hz for training. The policy receives the current robot pose 
and two RealSense camera images at $640{\times}360$, and encodes them using two pretrained ResNet-18 image encoders. It predicts a horizon of $32$ absolute joint-angle actions into the future. During execution, we only execute the first $16$ actions before re-planning. This follows the standard closed-loop policy execution setup, where executing a shorter prefix helps reduce compounding errors from stale predictions. 
The video generation model (Hunyuan1.5)~\citep{wu2025hunyuanvideo15technicalreport} is finetuned on the successful and failure demonstrations used to train the critic/policy, along with random trajectories. We finetune the model per-task across 8 A100 GPUs.

At each re-planning step, we sample 100 actions from the base policy. We perform greedy farthest point sampling in weighted joint space where we down weight the gripper contribution by 0.5 (as this often dominates the variance). Each of the $K=5$ remaining candidates is rolled forward with the action-conditioned video model, and we use the final frames of the generated videos as the input to the critic. The robot observes the new state and repeats the sample, generate, compare, and execute loop until task completion or a fixed max number of steps.

\textbf{Results.}
Table~\ref{tab:real_world_policy_results} reports real-world policy performance. On all evaluated tasks, our critic improves success rate over the base diffusion policy. We improve Stacking by +18\%, and Push Bowl experiments show a 2.3x improvement in mIoU. In the cluttered and difficult settings of Pickup Lego and Pick and Place Lego to bowl, our method improves success rate by +10\%/+4\%. These results suggest that strong critics can find the needle in the action haystack to improve performance of stochastic policies. For example, in the Lego task, different diffusion samples can lead to visibly different gripper-object alignments as seen in Figure~\ref{fig:realworld_qual_results} row 3, column b. The critic is able to select the image that shows a stable grasp between the robot gripper and the Lego. 

We also compare our critic against a critic trained without failure rollouts (Success-only), and find that similar to the baseline, the success-only critic achieves 12/50 on the Pickup Lego task and 6/50 on PickPlace Lego to Bowl. This supports our central hypothesis: for critic-in-the-loop policy improvement, it is not sufficient to recognize progress along successful demonstrations. The critic must also learn what failures look like to allow the policy to improve upon its mistakes.

Finally, we compare the average mean squared error (MSE) between the video generated by the candidate action that the critic chooses to execute and the ground truth observation video that actually rolls out in the real world upon executing $a*$. On cases where the policy successfully completes the task, the video model scores an average MSE of $0.00467$ across all generated videos along the execution. When the policy fails to complete the task, the video model has a higher MSE of $0.00523$. 
\begin{table*}[t]
\centering
\footnotesize
\setlength{\tabcolsep}{4pt}
\renewcommand{\arraystretch}{1.05}

\begin{tabular}{@{}lcc@{}}
\toprule
& \multicolumn{2}{c}{\textbf{Diffusion Policy}}\\
\cmidrule(lr){2-3}
\textbf{Task} & \textbf{Base} & \textbf{Ours}\\
\midrule
Stacking               & 23 / 50    & \textbf{32 / 50}\\
Push-Bowl              & 0.140 mIoU & \textbf{0.321 mIoU}\\
Pickup Lego            & 11 / 50    & \textbf{16 / 50}\\
PickPlace Lego To Bowl & 6 / 50     & \textbf{8 / 50}\\
\bottomrule
\end{tabular}

\vspace{2pt}
\textbf{(a) Real World}

\vspace{12pt}

\scriptsize
\begin{tabular}{@{}lcc@{\hspace{1.6em}}lcc@{}}
\toprule
& \multicolumn{2}{c}{\textbf{GR00T N1.5}} & & \multicolumn{2}{c}{\textbf{GR00T N1.5}}\\
\cmidrule(lr){2-3}\cmidrule(lr){5-6}
\textbf{Task} & \textbf{Base} & \textbf{Ours} & \textbf{Task} & \textbf{Base} & \textbf{Ours}\\
\midrule
NavigateKitchen           & $0.040 \pm 0.009$          & $0.040 \pm 0.009$          & PickPlaceCounterToStove   & $\mathbf{0.696 \pm 0.015}$ & $0.568 \pm 0.020$\\
CloseBlenderLid           & $\mathbf{0.072 \pm 0.016}$ & $0.013 \pm 0.007$          & PickPlaceDrawerToCounter  & $0.116 \pm 0.021$          & $\mathbf{0.180 \pm 0.021}$\\
CloseFridge               & $0.772 \pm 0.031$          & $\mathbf{0.863 \pm 0.014}$ & PickPlaceSinkToCounter    & $\mathbf{0.888 \pm 0.022}$ & $0.757 \pm 0.022$\\
CloseToasterOvenDoor      & $0.356 \pm 0.017$          & $\mathbf{0.623 \pm 0.032}$ & PickPlaceToasterToCounter & $\mathbf{0.780 \pm 0.019}$ & $0.729 \pm 0.022$\\
CoffeeSetupMug            & $0.160 \pm 0.028$          & $0.184 \pm 0.031$          & SlideDishwasherRack       & $0.300 \pm 0.021$          & $\mathbf{0.530 \pm 0.018}$\\
OpenCabinet               & $0.500 \pm 0.025$          & $\mathbf{0.552 \pm 0.019}$ & TurnOffStove              & $0.104 \pm 0.021$          & $\mathbf{0.203 \pm 0.017}$\\
OpenDrawer                & $0.544 \pm 0.010$          & $\mathbf{0.648 \pm 0.019}$ & TurnOnElectricKettle      & $0.348 \pm 0.016$          & $\mathbf{0.620 \pm 0.032}$\\
OpenStandMixerHead        & $0.588 \pm 0.023$          & $\mathbf{0.713 \pm 0.015}$ & TurnOnMicrowave           & $0.384 \pm 0.025$          & $\mathbf{0.427 \pm 0.029}$\\
PickPlaceCounterToCabinet & $0.760 \pm 0.017$          & $0.752 \pm 0.030$          & TurnOnSinkFaucet          & $0.440 \pm 0.017$          & $\mathbf{0.500 \pm 0.013}$\\
\midrule
\multicolumn{6}{r}{\textbf{Average}\hspace{2em}Base: $0.436 \pm 0.020$\hspace{2em}Ours: $\mathbf{0.495 \pm 0.021}$}\\
\bottomrule
\end{tabular}

\vspace{2pt}
\normalsize\textbf{(b) RoboCasa365}

\caption{\textbf{Policy results.} We compare baseline Diffusion Policy and GR00T N1.5 against our critic-guided policies on four real-world and eighteen RoboCasa365 simulation atomic tasks. For RoboCasa365, we report mean success rate $\pm$ standard error across 5 runs. The critic selects candidate action proposals from the base policy that help the policy avoid failures, leading to overall improvement in task success rate.}
\label{tab:real_world_policy_results}
\end{table*}

\subsection{Simulation Experimental Setup and Results}
\label{sec:sim_policy_results}
\textbf{Setup.} Alongside real world experiments, we conduct experiments on the 18 atomic tasks from the official RoboCasa365 simulation framework~\citep{nasiriany2026robocasa365largescalesimulationframework}. Our base policy $\pi$ is a multi-task language-conditioned policy. We use the publicly released Robocasa365 checkpoint which starts from pretrained checkpoints released by NVIDIA Isaac GR00T N1.5~\citep{gr00tn1_2025}.  This multi-task policy is trained on data across 300 tasks, comprising 65 atomic tasks and 235 composite tasks. For each task, 100 task demonstrations per task are used, resulting in 482 hours of total data. 

We follow the Robocasa365 setup by evaluating on the pretrain scenes. To gather failure demonstrations,
we rollout GR00T N1.5 on 50 different initial conditions across 5 different seeds from the pretraining set, following the protocol in Robocasa365 benchmarking. Of these 250 rollouts, the policy exhibits variance, creating a dataset of failures to train the VLM critic. The breakdown of total (train and validation) number of demonstrations and number
of success/failure pairs before balancing are provided in Appendix Table \ref{tab:vlm_dataset_details}.  We follow the same evaluation parameters with an action horizon of 16 and standard closed-loop policy execution setup. We train a multi-task critic with the same procedure as in real. Candidate videos are rendered by the simulator for these tasks. We use the same hyperparameters as real for critic-in-the-loop with N=100, K=5, and greedy farthest point sampling. 

\textbf{Results.} As seen in Table~\ref{tab:real_world_policy_results}, across all tasks, our critic in-the-loop achieves 49.5\% success rate compared to baseline GR00T N1.5 of 43.6\%, an average gain of +5.9\%. Our method shows significant improvements in tasks that require pushing/pulling buttons, doors, and levers. 

\subsection{Limitations}

Weak action-conditioned video model predictions can cause failures. For example, we sometimes see the model hallucinate grasping the Lego in the predicted video when the real-world rollout fails to grasp, or the bowl pushed to a slightly different end location than the ground truth. We expect these errors to reduce as video models improve. The video model takes $\sim$25 seconds to synthesize five videos in parallel (across 8 A100 GPUs and 10 inference steps), and the critic takes $\sim$1 second to generate all pair-wise rankings in parallel (across 8 A100 GPUs). We expect the video model's latency to reduce as faster implementations are developed.
\section{Conclusion}
\label{sec:conclusion}

We studied a method to finetune VLM critics for pairwise progress/failure detection by using success and failure rollouts. The resulting critic substantially improves offline accuracy in simulation and on a real-world rollout dataset, particularly on success–failure discrimination. We put these critics into closed-loop policies to select between several candidate future states generated from an action-conditioned video model, improving average policy success rate by 11\% across real-world tasks and 5.9\% across simulation tasks.

\clearpage
\acknowledgments{We thank Matei Ciocarlie and Huy Ha for key discussions. This research is based on work partially supported by the Toyota Research Institute and NSF awards \#2046910 and \#2132519. This work is also partially supported by the funds provided by the National Science Foundation and by DoD OUSD (R\&E) under Cooperative Agreement PHY-2229929 (The NSF AI Institute for Artificial and Natural Intelligence).}

\bibliography{main}  
\clearpage
\clearpage
\appendix
\section*{Appendix}
\section{Additional Critic results}

\subsection{Fine vs coarse performance}
\label{sec:finevscoarse}
\begin{figure}[H]
    \centering
    \captionsetup{skip=5pt} 
    \begin{subfigure}{0.49\textwidth}
        \centering
        \includegraphics[width=\linewidth]{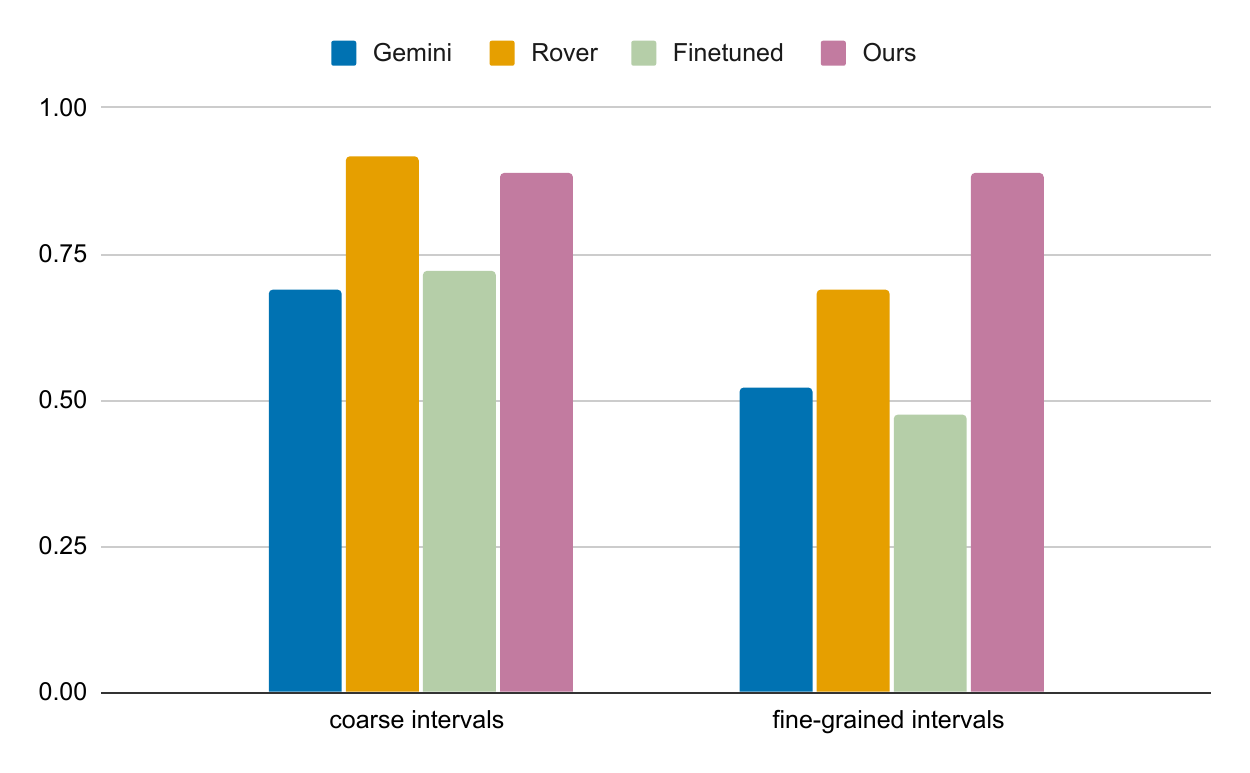}
        \caption{RoboCasa}
        \label{fig:robocasa_frame_interval}
    \end{subfigure}
    \hfill
    \begin{subfigure}{0.49\textwidth}
        \centering
        \includegraphics[width=\linewidth]{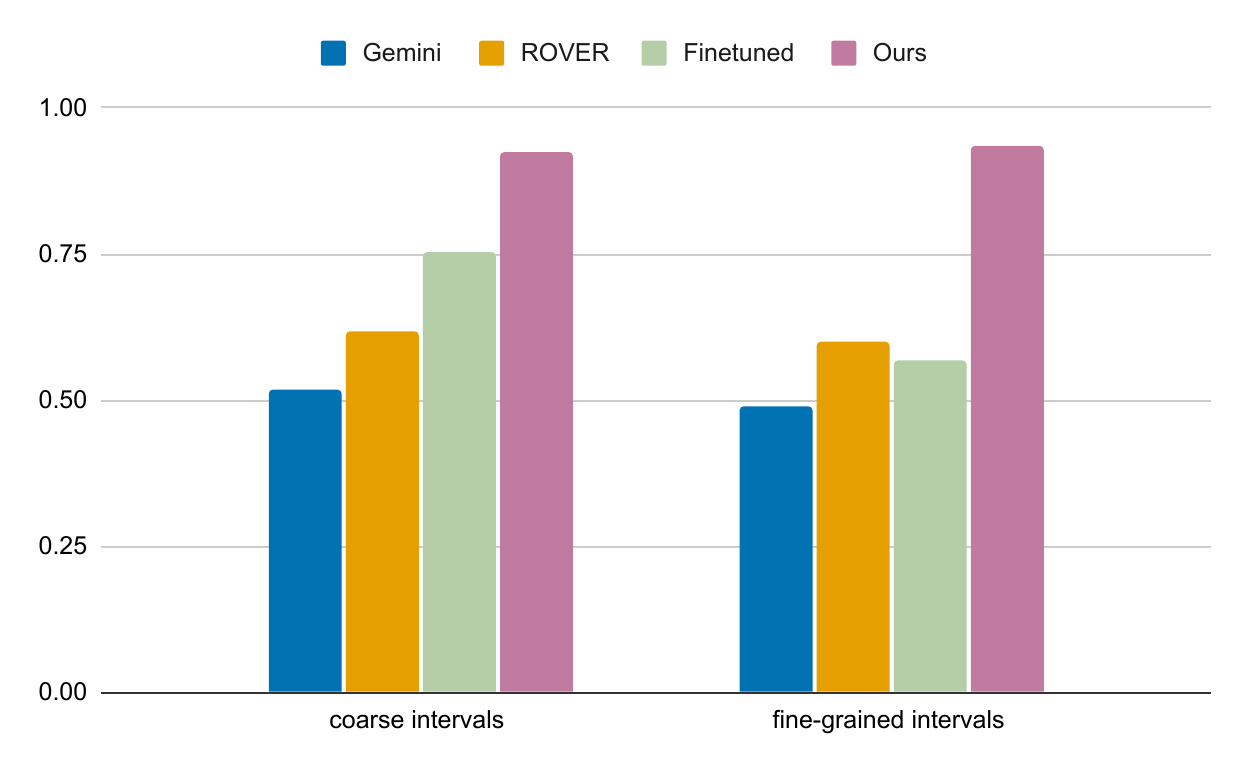}
        \caption{Real World}
        \label{fig:realworld_frame_interval}
    \end{subfigure}
    \caption{\textbf{Coarse vs fine-grained critic performance}. Current VLM performance degrades when judging fine grained visual differences in task performance. Fine-tuning for fine-grained task progress recognition can boost these scores.}
    \label{fig:intervals}
\end{figure}

To isolate the effect of visual difference magnitude, we evaluate ROVER, ProgressLM, and our method under two sampling regimes. In the \textbf{fine-grained} regime, we sample query frames with small temporal intervals ($l=16,32,48$ steps apart), where differences are subtle. In the \textbf{coarse} regime, we sample frames at larger intervals ($l=74,100,150$), where more pronounced changes occur. Figure~\ref{fig:robocasa_frame_interval} and Figure~\ref{fig:realworld_frame_interval} show that increasing the frame interval substantially boosts performance for prompted models, indicating that these models are more reliable when differences are large. However, practical policy guidance requires performance in the fine-grained regime as well.

\subsection{Generalization to unseen tasks}
\label{sec:gen_unseen_tasks}
We evaluate the three strongest methods (ROVER, ProgressLM, and ours) on held-out tasks that are unseen for ProgressLM and our method. ROVER remains prompted and continues to receive one-shot examples, as in other splits. Our method generalizes to new tasks that it was never fine-tuned on (Table~\ref{tab:newtaskstable}), suggesting a promising path toward broader generalization as training data and task diversity scale. We believe that this VLM critic finetuning recipe can build strong critic backbones with task finetuning for more learning about policy specific success/failures. 
\begin{table}[H]
\centering
\scriptsize
\setlength{\tabcolsep}{3.2pt}
\renewcommand{\arraystretch}{1.08}
\resizebox{0.8\textwidth}{!}{%
\begin{tabular}{l | c c c}
\toprule
\textbf{Task}
& \textbf{ROVER}
& \textbf{ProgressLM}
& \textbf{Ours (Success+Failure)} \\
\midrule
TurnSinkSpout & 0.372 & 0.601 & 0.700 \\
OpenDrawer & 0.559 & 0.565 & 0.825 \\
CloseSingleDoor & 0.490 & 0.365 & 0.750 \\
CloseDoubleDoor & 0.571 & 0.583 & 0.695 \\
OpenDoubleDoor & 0.740 & 0.783 & 0.655 \\
OpenSingleDoor & 0.473 & 0.652 & 0.815 \\
\midrule
\textbf{Avg. Multi-task Acc.} 
& \textbf{0.533}
& \textbf{0.591}
& \textbf{0.740} \\
\bottomrule
\end{tabular}%
}
\caption{\textbf{Generalization to tasks unseen} during finetuning for ProgressLM and our method (ROVER remains prompted with one-shot examples). Our method generalizes to new tasks never seen during finetuning.}
\label{tab:newtaskstable}
\end{table}

\section{Additional Policy results}
\label{sec:accvsk_results}
\begin{figure}[!htbp]
    \centering
    \captionsetup{skip=4pt}

    \begin{minipage}[t]{0.46\linewidth}
        \centering
        \includegraphics[width=\linewidth]{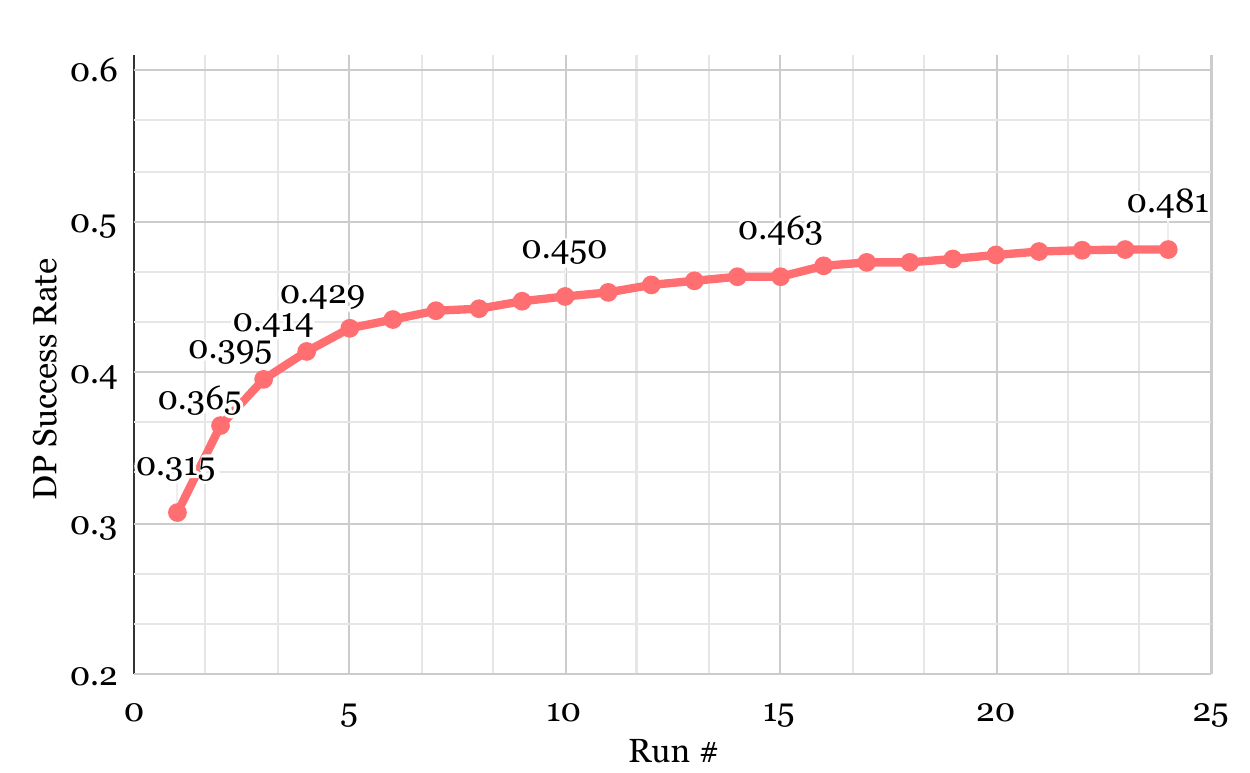}
        \caption{\textbf{Best-of-$K$ sampling} reveals substantial headroom for Diffusion Policy, increasing success rate from 31\% to 48\%.}
        \label{fig:oracle_gain_dp}
    \end{minipage}
    \hfill
    \begin{minipage}[t]{0.53\linewidth}
        \centering
        \includegraphics[width=\linewidth]{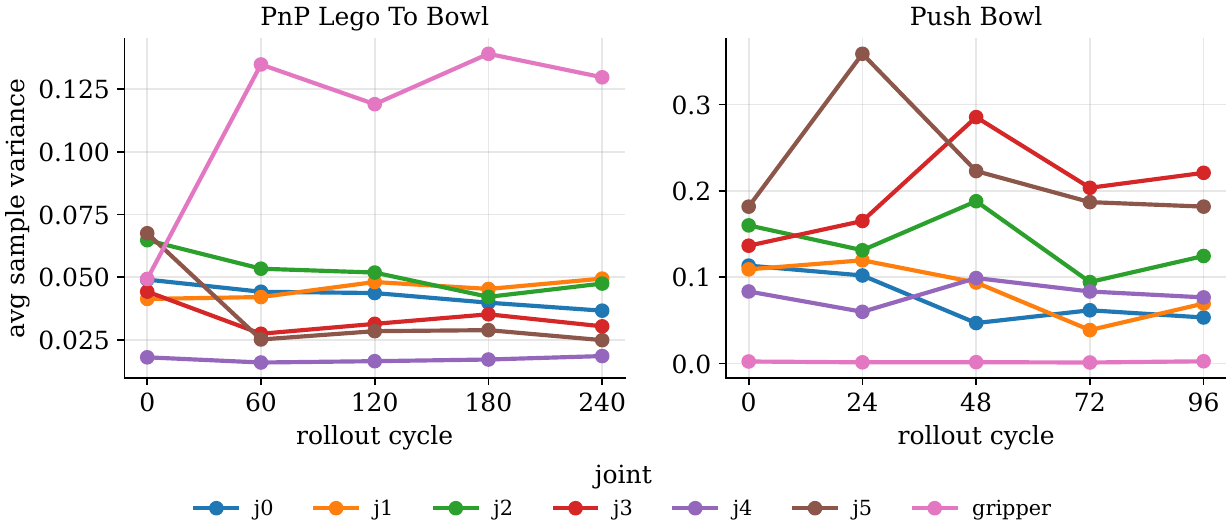}
        \caption{\textbf{Per-joint sample variance} across K=5 samples. Variance indicates the multimodality of the policy.}
        \label{fig:policy_variance}
    \end{minipage}

    \vspace{-0.15in}
\end{figure}

Strong robot policies are typically represented as stochastic distributions, and most behavior cloning evaluations report the mean success rate over a small number of independent runs (different random seeds). Figure~\ref{fig:oracle_gain_dp} shows that rather than averaging, best-of-$K$ sampling yields large gains even for $K \le 5$, indicating that successful action sequences frequently exist in the policy's support but are not sampled reliably under a single rollout. Furthermore, Figure~\ref{fig:policy_variance} shows the variance in joint values output by the policy when taking 5 samples at test time. This motivates using a learned critic to select among candidate rollouts at test time. 

\section{Experiment Details}
\subsection{Experimental Setup}
We set up two stereo cameras (Intel RealSense D435i cameras) spaced approximately 660 mm apart at a 45 degree angle. The distance between the cameras and the table is about 500 mm. The realworld demonstrations are recorded at a resolution of 1280×720 (each model crops and resizes it to fit the specific model's input requirements). The robot used is I2RT's Yam Pro Arm with 7dof, and we record/store the absolute joint angles of the follower. We use the Yam Pro Leader Arm for teleoperation during data collection. For all tasks, we allow the policy to rollout for a maximum number of steps which is set to 1.5x the average number of steps used during data collection. Task success for the Pick and Place Lego to Bowl and Stacking tasks are binary and defined by a human. For calculating the mIoU in the Push Bowl experiment, the view from one of the stereo cameras is used. 

\label{sec:dataset_details}
\begin{table}[!htbp]
\centering
\footnotesize
\setlength{\tabcolsep}{6pt}
\renewcommand{\arraystretch}{1.05}
\begin{tabular}{llrr}
\toprule
\textbf{Task} & \textbf{Type} & \textbf{\# Demos} & \textbf{Avg. Length} \\
\midrule
\multirow{4}{*}{Push-Bowl}
    & Success & 30 & 227.7 \\
    & Failure & 11 & 367.0 \\
    & Random  & 10 & 4,359.9 \\
\midrule
\multirow{4}{*}{Stacking}
    & Success & 50 & 184.5 \\
    & Failure & 39 & 187.3 \\
    & Random  & 20 & 825.5 \\
\midrule
\multirow{4}{*}{PickPlace Lego to Bowl}
    & Success & 200 & 265.6 \\
    & Failure & 49 & 760.7 \\
    & Random  & 10 & 2,675.4 \\
\bottomrule
\end{tabular}
\vspace{0.6em}
\caption{\textbf{Real World dataset} statistics by task and trajectory type. Average length is measured in timesteps per demonstration.}
\label{tab:realworld_dataset_stats}
\end{table}

\begin{table*}[!tbp]
\centering
\footnotesize

\begin{minipage}[t]{0.48\textwidth}
\centering
\begin{tabular}{lrrrr}
\toprule
Task & \# Succ. & \# Fail & \# S--S & \# S--F\\
\midrule
Lego (PnP) & 69 & 39 & 10,829 & 1,438 \\
Push Bowl  & 30 & 11 & 14,376 & 1,446 \\
Stacking   & 48 & 38 & 18,890 & 3,134 \\
\bottomrule
\end{tabular}

\vspace{2mm}
{\small (a) Real-world tasks}
\end{minipage}
\hfill
\begin{minipage}[t]{0.48\textwidth}
\centering
\begin{tabular}{lrrrr}
\toprule
Task & \# Succ. & \# Fail & \# S--S & \# S--F \\
\midrule
 CloseBlenderLid & 18 & 232 & 836 & 474 \\
 CloseFridge & 193 & 57 & 40,934 & 3,887 \\
 CloseToasterOvenDoor & 89 & 161 & 7,895 & 3,616 \\
 CoffeeSetupMug & 40 & 210 & 4,271 & 2,754 \\
 OpenCabinet & 125 & 125 & 20,235 & 5,595 \\
 OpenDrawer & 136 & 114 & 13,175 & 3,712 \\
 OpenStandMixerHead & 147 & 103 & 8,206 & 1,274 \\
 PickPlaceCounterToCabinet & 190 & 60 & 19,809 & 1,071 \\
 PickPlaceCounterToStove & 174 & 76 & 19,894 & 2,277 \\
 PickPlaceDrawerToCounter & 29 & 221 & 2,896 & 1,832 \\
 PickPlaceSinkToCounter & 222 & 28 & 26,865 & 880 \\
 PickPlaceToasterToCounter & 195 & 55 & 28,107 & 3,143 \\
 SlideDishwasherRack & 75 & 175 & 5,088 & 1,339 \\
 TurnOffStove & 26 & 224 & 2,729 & 1,938 \\
 TurnOnElectricKettle & 87 & 163 & 6,077 & 2,231 \\
 TurnOnMicrowave & 96 & 154 & 10,505 & 5,370 \\
 TurnOnSinkFaucet & 110 & 140 & 12,251 & 6,889 \\
\bottomrule
\end{tabular}

\vspace{2mm}
{\small (b) RoboCasa-365 tasks}
\end{minipage}

\caption{\textbf{VLM critic finetuning datasets}. For real-world tasks, we use human collected data demonstrating success and failures. For Robocasa365, we rollout on 50 initial conditions per task 5 times, which generates multiple success/failure demos (counts shown in columns). From these gathered demos, we generate success--success (S--S) and success--failure (S--F) pairs using our finetuning method. Raw counts are shown before balancing.}
\label{tab:vlm_dataset_details}
\end{table*}            

\begin{wraptable}[11]{r}{0pt}
\vspace{-0.35in}
\centering
\tiny
\setlength{\tabcolsep}{2.5pt}
\renewcommand{\arraystretch}{0.78}
\begin{tabular}{@{}lclc@{}}
\toprule
Hyperparam. & Value & Hyperparam. & Value \\
\midrule
State Norm. & Yes & Action Norm. & Yes \\
Img. Chunk & 2 & Img. Size & $640{\times}360$ \\
State Dim. & 7 & Action Dim. & 7 \\
Exec. Hor. & 16 & Pred. Hor. & 32 \\
Batch Size & 64 & Epochs & 250 \\
LR & $10^{-4}$ & Workers & 4 \\
Train Steps & 100 & Infer. Steps & 16 \\
\bottomrule
\end{tabular}

\vspace{-0.08in}
\caption{\textbf{Diffusion Policy training hyperparameters} for real-world experiments.}
\label{tab:base_policy_hyperparameters}
\vspace{-0.16in}
\end{wraptable}
\subsection{Base Policy }
We use Diffusion Policy~\citep{Chi2023DiffusionPV} as our robot policy for real world experiments which has the ability to model multimodal distributions and complex robot behaviors. We train a per-task policy on all successful demos for real-world. In simulation, we use the finetune checkpoint GR00T N1.5~\cite{gr00tn1_2025} which is a single language-conditioned multi-task policy. Training hyperparameter details are listed in Table~\ref{tab:base_policy_hyperparameters}. 

\subsection{Action-Conditioned Video Generation Model}
\label{sec:video_model_details}
We finetune the pretrained HunyuanVideo-1.5 model~\citep{wu2025hunyuanvideo15technicalreport} by adapting its architecture to incorporate action conditioning. Specifically, we add cross-attention blocks to the video generation Diffusion Transformer (DiT), enabling the model to condition video synthesis on robot action trajectories.

We denote the model input as $(x_0, a_{0:T})$, where $x_0 \in \mathbb{R}^{H \times W \times 3}$ is the initial RGB image observation of the environment and $a_{0:T} \in \mathbb{R}^{T \times 7}$ is the sequence of 7-DoF robot arm states corresponding to the $T$ frames of the generated video. For action conditioning, each robot arm state is projected into a single action token with dimension $D_a$ using an input MLP, producing an action-token sequence in $\mathbb{R}^{T \times D_a}$. A 4-layer transformer encoder then processes these tokens along the temporal dimension while preserving the sequence shape, yielding encoded action tokens in $\mathbb{R}^{T \times D_a}$. The resulting action tokens are used as conditioning inputs to cross-attention blocks inserted before the self-attention block in each layer of the video generation DiT, and the video tokens query the encoded action-token sequence through cross-attention. The outputs of the cross-attention blocks are further passed through a projection layer to produce action-dependent modulation signals. These signals are injected into the DiT through the timestep modulation pathway by adding the action modulation to the original timestep modulation. We initialize the model from the pretrained checkpoint and finetune it on all demonstration data described in Table~\ref{tab:realworld_dataset_stats}. Training uses the standard flow-matching video generation objective adopted during pretraining of the base video model~\citep{wu2025hunyuanvideo15technicalreport}. The model and training hyperparameters are summarized in Table~\ref{tab:world_model_hyperparameters}.

For the action-conditioned video generation model, we found that sufficient diversity in the random demonstrations is important for enabling the world model to learn a broad range of environment state transitions.

\begin{table}[!htbp]
\centering
\begin{tabular}{@{}lc@{}}
\toprule
\textbf{Hyperparameter} & \textbf{Value} \\
\midrule
Base Model & Pretrained HunyuanVideo-1.5 \\
LoRA & Rank 64 on Pretrained Blocks \\
Training Precision & bf16 \\
Video Resolution & 848$\times$480 \\
Video Length & 33 frames \\
Action Length & 33 frames \\
Optimizer & Muon \\
Action Blocks Learning Rate & $1e-4$ \\
LoRA Learning Rate & $1e-5$ \\
Weight Decay & 0.01 \\
Effective Batch Size & 32 \\
Training Steps & 4000 \\
\bottomrule
\end{tabular}
\vspace{0.6em}
\caption{\textbf{Action-conditioned video generation model hyperparameters.}}
\label{tab:world_model_hyperparameters}
\end{table}

\subsection{VLM critic}
\label{sec:prompt_details}
\begin{figure}[!htbp]
    \centering
    \begin{subfigure}[t]{0.48\linewidth}
        \centering
        \includegraphics[width=\linewidth]{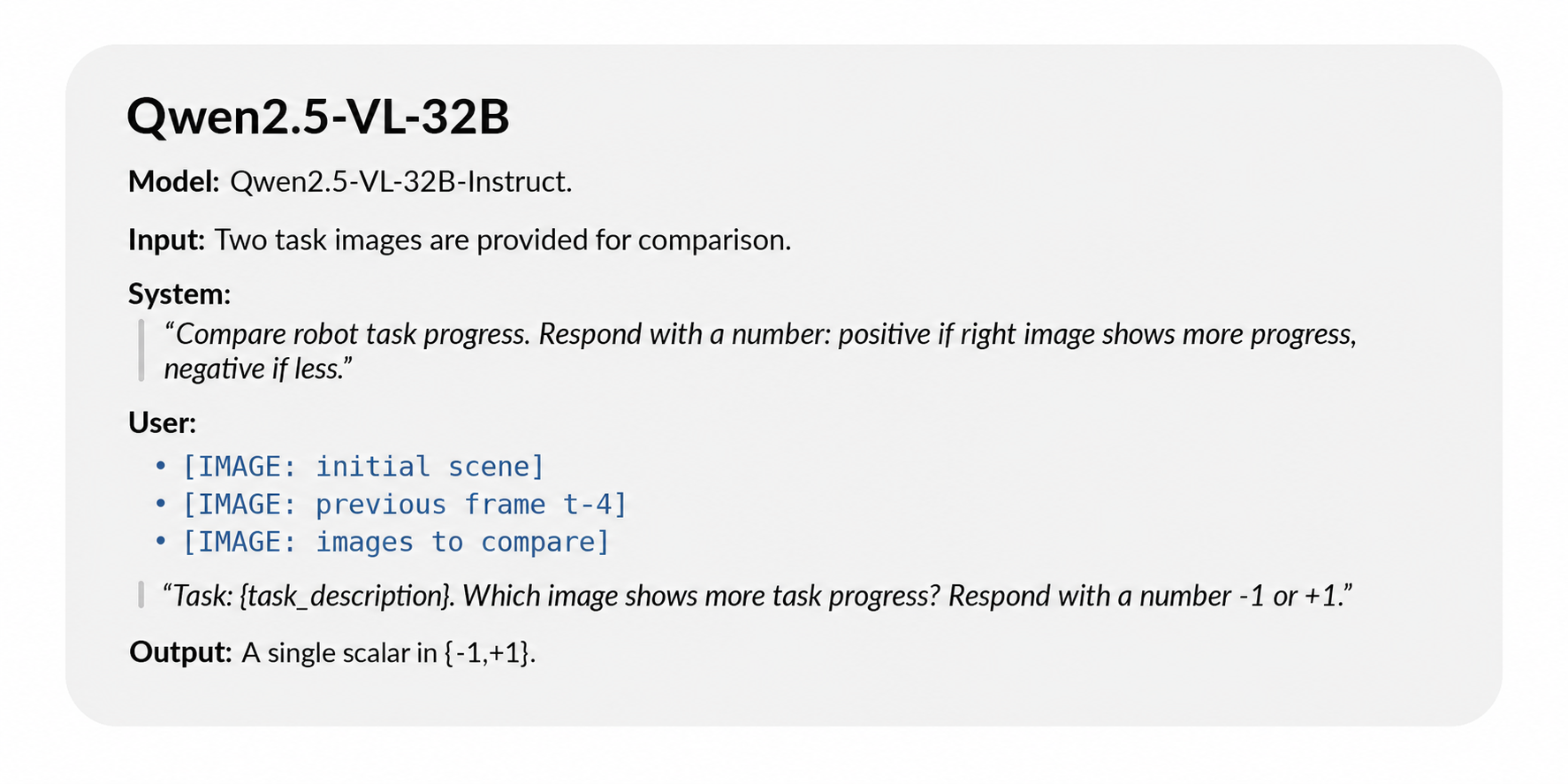}
    \end{subfigure}
    \hfill
    \begin{subfigure}[t]{0.48\linewidth}
        \centering
        \includegraphics[width=\linewidth]{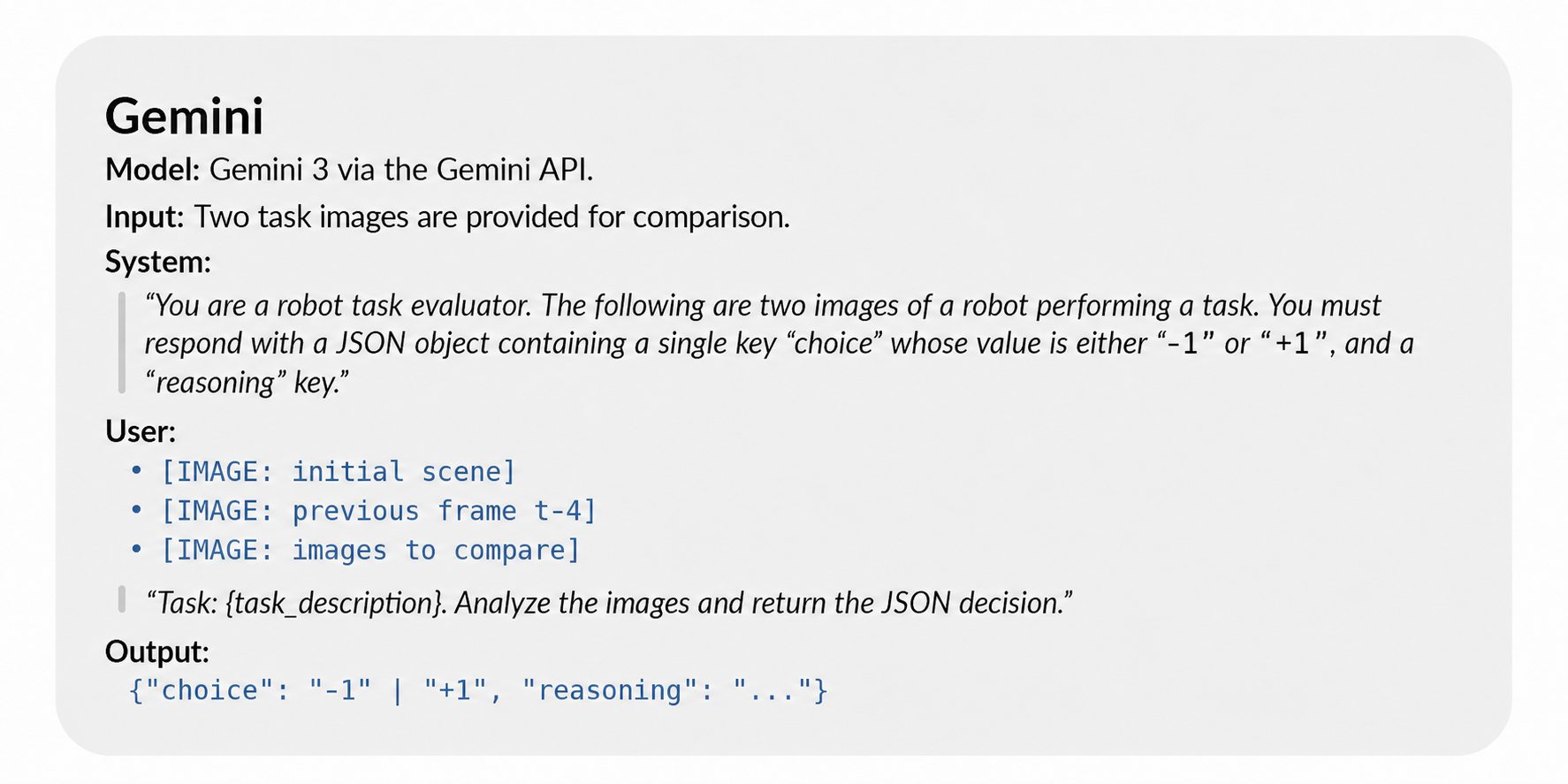}
    \end{subfigure}
    \vspace{0.6em}
    \begin{subfigure}[t]{0.48\linewidth}
        \centering
        \includegraphics[width=\linewidth]{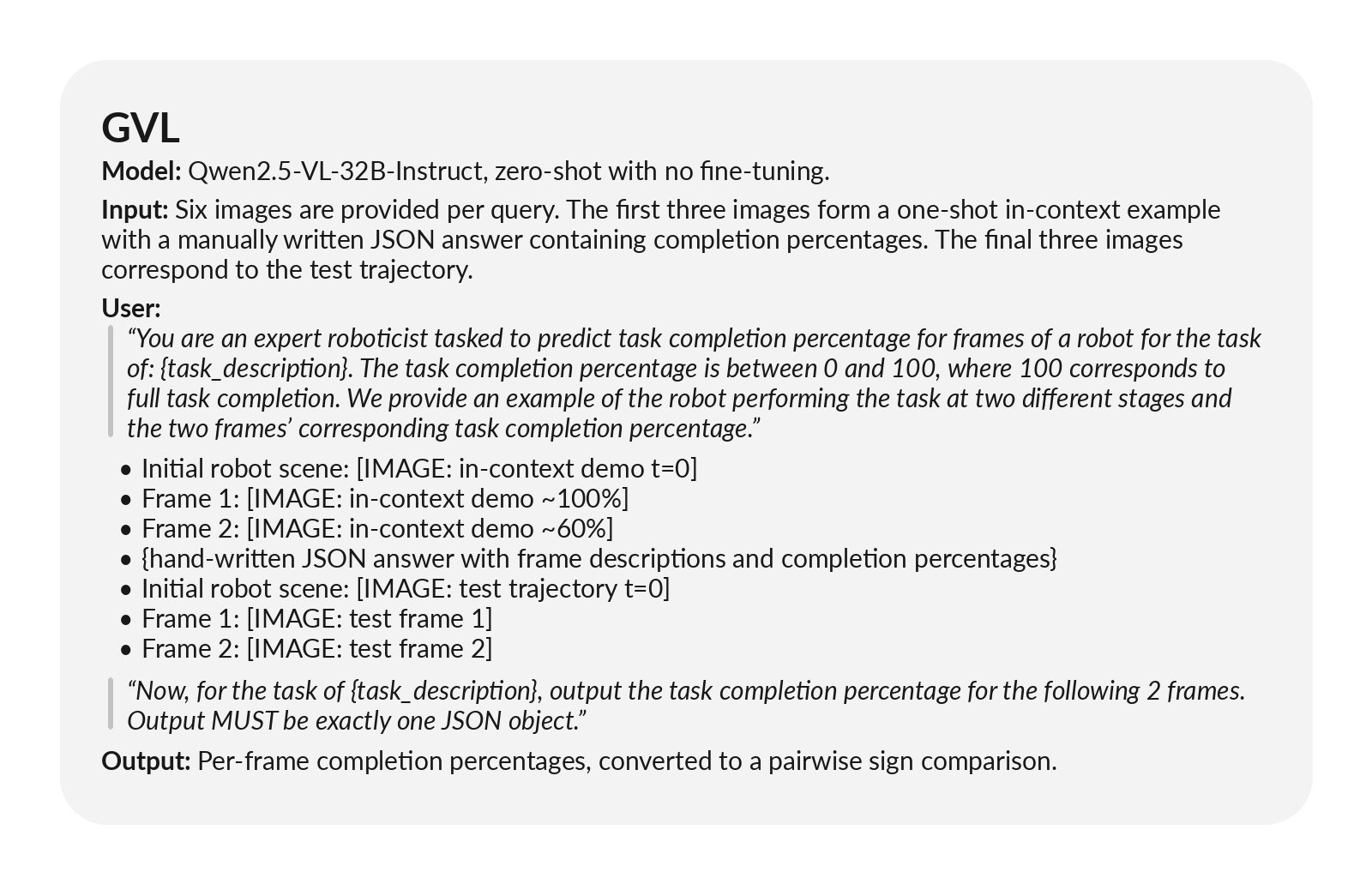}
    \end{subfigure}
    \hfill
    \begin{subfigure}[t]{0.48\linewidth}
        \centering
        \includegraphics[width=\linewidth]{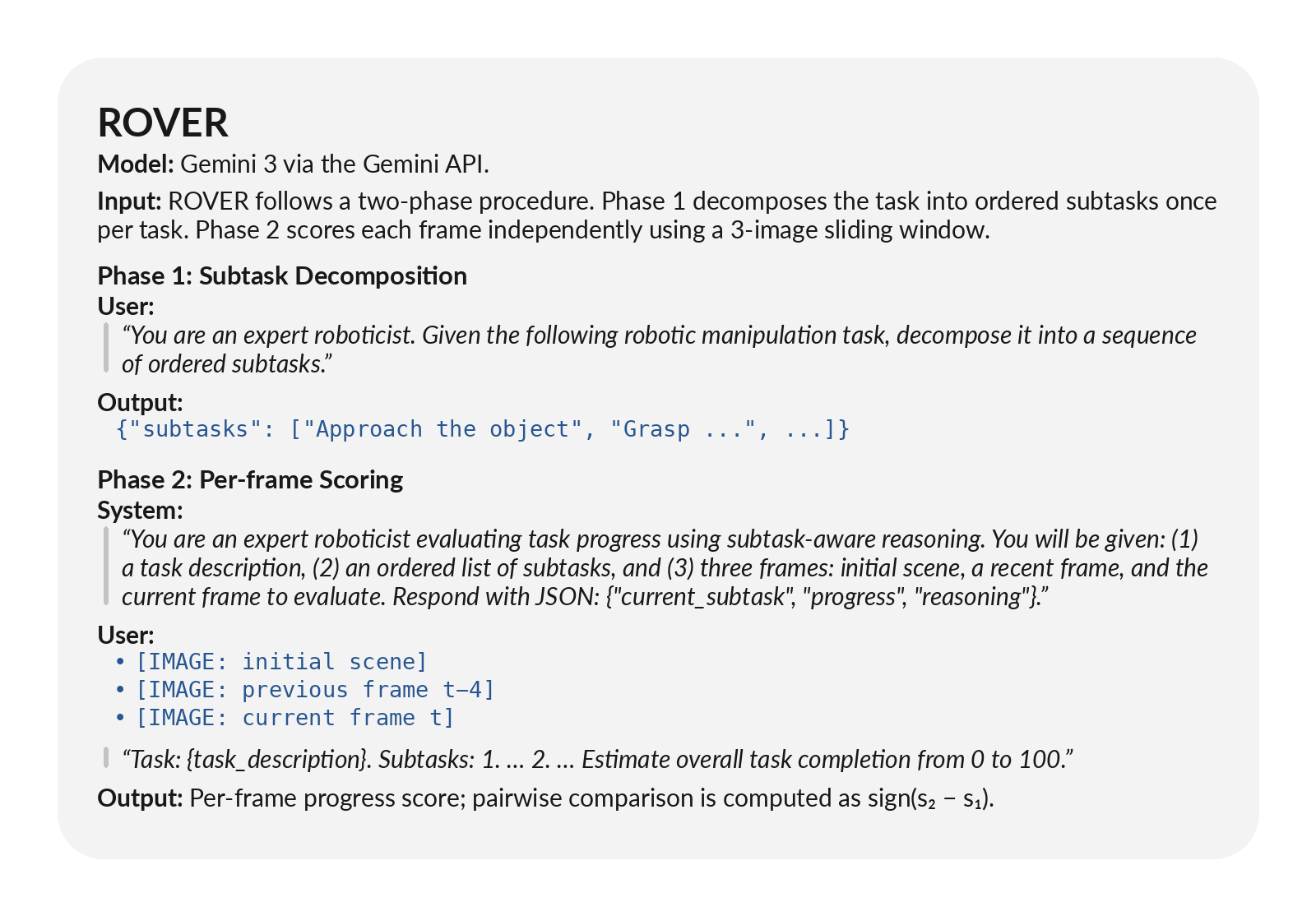}
    \end{subfigure}
    \begin{subfigure}[t]{0.48\linewidth}
        \centering
        \includegraphics[width=\linewidth]{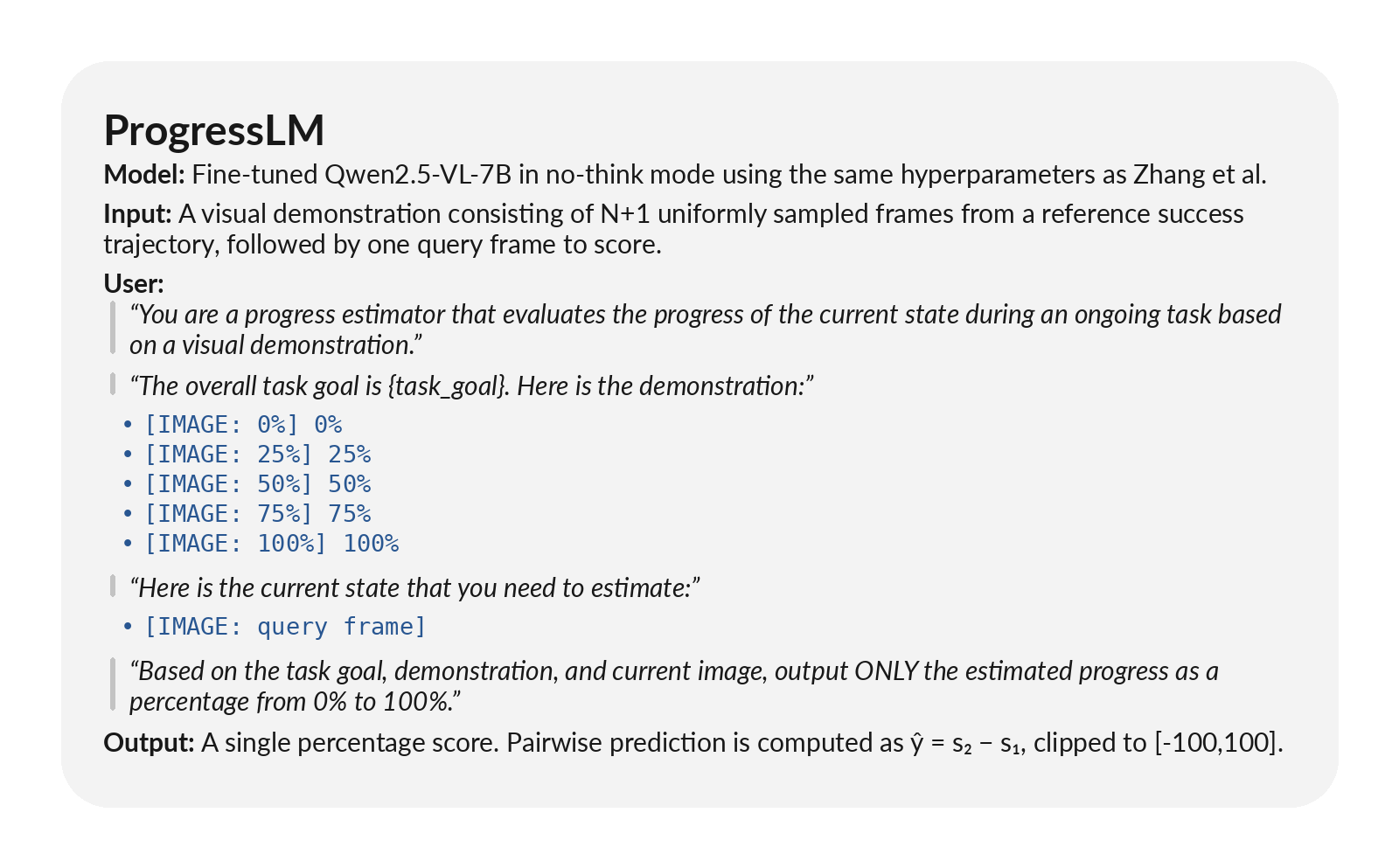}
    \end{subfigure}
    \hfill
    \begin{subfigure}[t]{0.48\linewidth}
        \centering
        \includegraphics[width=\linewidth]{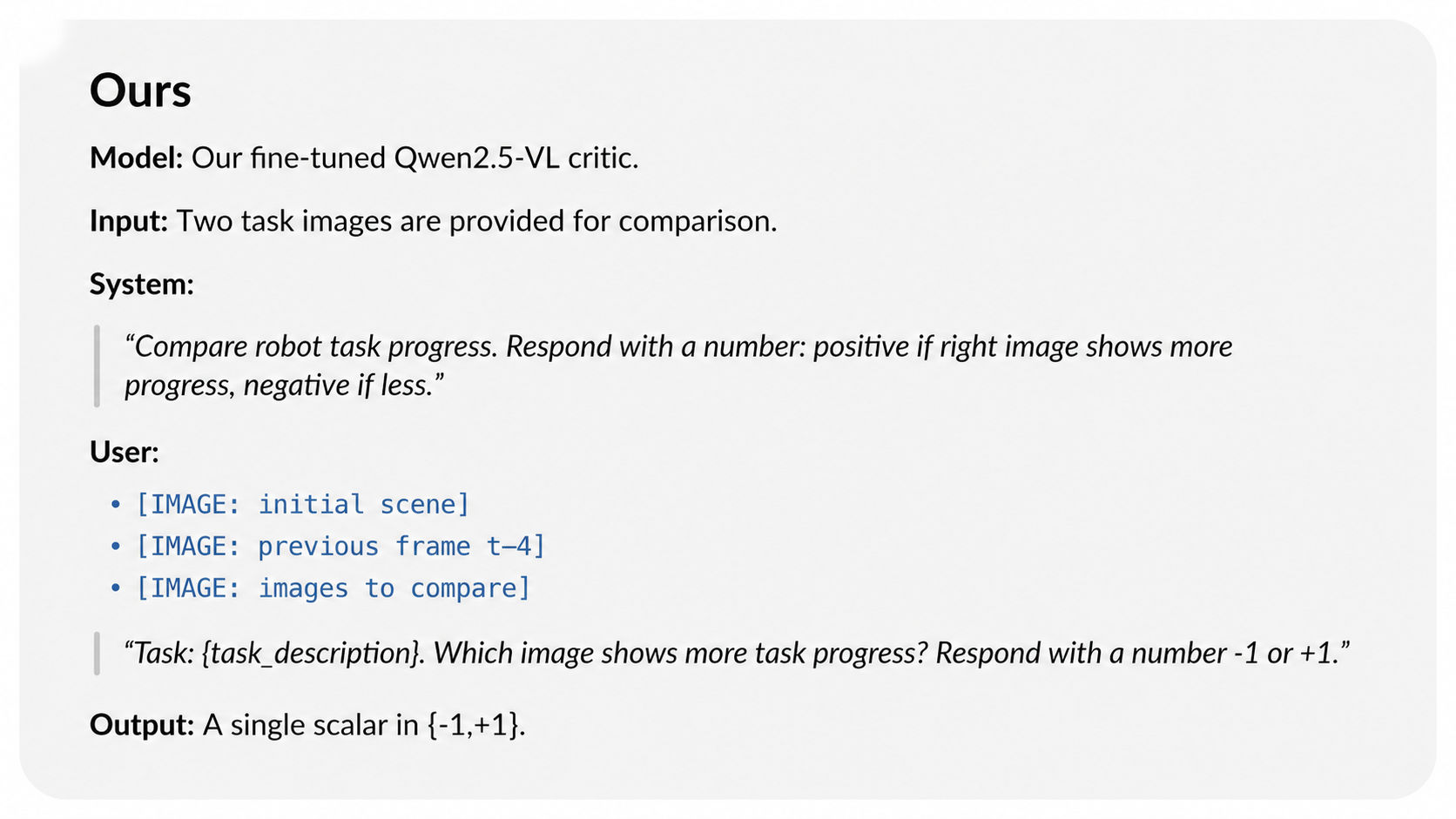}
    \end{subfigure}

    \caption{\textbf{Prompt templates} used for baselines and our method.}
    \label{fig:prompt_templates}
\end{figure}

We finetune the VLM critic on the collected successful and failure data (dataset details provided in Table~\ref{tab:vlm_dataset_details}). The critic in the real world is trained per-task while for simulation we train a single multi-task critic across all 18 atomic tasks. We provide the details of the hyperparameters used to finetune the VLM critic in Table~\ref{tab:vlm_hyperparameters}. Furthermore, the details on the prompts used per baseline in Figure \ref{fig:prompt_templates}.
  \begin{table}[!htbp]
  \centering
  \scriptsize
  \renewcommand{\arraystretch}{1.03}
  \setlength{\tabcolsep}{3pt}

  \begin{minipage}[t]{0.49\linewidth}
  \centering
  \resizebox{\linewidth}{!}{%
  \begin{tabular}{@{}lc@{}}  
  \toprule  
  \textbf{Hyperparameter} & \textbf{Value} \\
  \midrule  
  Base Model & Qwen2.5-VL-3B-Instruct \\     
  Fine-tuning & Full (all parameters) \\     
  Attention Implementation & FlashAttention-2 \\
  Mixed Precision & bf16 \\  
  Distributed Backend & DeepSpeed ZeRO-2 \\  
  Number of GPUs & 8 \\      
  Images per Comparison & 2 \\               
  Max Image Resolution & 960$\times$540 \\   
  Optimizer & AdamW \\       
  Learning Rate & $1e-5$ \\  
  LR Schedule & Linear \\    
  Warmup Ratio & 0.05 \\     
  Weight Decay & 0.0 \\      
  Max Gradient Norm & 1.0 \\ 
  Per-Device Batch Size & 4 \\               
  Gradient Accumulation Steps & 2 \\
  Effective Batch Size & 64 \\
  Number Steps & 1400 \\                   
  Completion-Only Loss & Yes \\                                              Balance Failure vs.\ Success & Yes \\
  \bottomrule
  \end{tabular}
  }
  \vspace{0.15em}
  {\footnotesize \textbf{(a)} Real World}
  \end{minipage}
  \hspace{0.01\linewidth}     
  \begin{minipage}[t]{0.49\linewidth}       
  \centering                  
  \resizebox{\linewidth}{!}{%
  \begin{tabular}{@{}lc@{}}   
  \toprule                                  
  \textbf{Hyperparameter} & \textbf{Value} \\
  \midrule                    
  Base Model & Qwen2.5-VL-3B-Instruct \\
  Fine-tuning & Full (all parameters) \\
  Attention Implementation & FlashAttention-2 \\
  Mixed Precision & bf16 \\
  Distributed Backend & DeepSpeed ZeRO-2 \\
  Number of GPUs & 8 \\                     
  Images per Comparison & 2 \\
  Max Image Resolution & 848$\times$480 \\
  Optimizer & AdamW \\        
  Learning Rate & $1.5e-5$ \\                 
  LR Schedule & Linear \\     
  Warmup Steps & 200 \\
  Weight Decay & 0.0 \\       
  Max Gradient Norm & 1.0 \\                
  Per-Device Batch Size & 12 \\
  Gradient Accumulation Steps & 2 \\
  Effective Batch Size & 192 \\
  Number Steps & 4500 \\                   
  Completion-Only Loss & Yes \\
  Balance Failure vs.\ Success & Yes \\                                                
  \bottomrule
  \end{tabular}}                                         
  \vspace{0.15em}

  {\footnotesize \textbf{(b)} RoboCasa365 Simulation}
  \end{minipage}

  \vspace{0.4em}
  \caption{\textbf{VLM Critic Fine-tuning Hyperparameters.}}
  \label{tab:vlm_hyperparameters}
  \end{table}

\end{document}